\documentclass{article}
\usepackage[accepted]{icml2024}

\usepackage{microtype}
\usepackage{graphicx}
\usepackage{subfigure}
\usepackage{booktabs} 

\usepackage{hyperref}

\usepackage{amsmath}
\usepackage{amssymb}
\usepackage{mathtools}
\usepackage{amsthm}

\usepackage{relsize}
\usepackage{bm}

\usepackage[textsize=tiny]{todonotes}

\theoremstyle{plain}
\newtheorem{theorem}{Theorem}

\theoremstyle{definition}
\newtheorem{definition}{Definition}

\theoremstyle{remark}

\begin{document}

\twocolumn[
\icmltitle{Hidden Synergy: $L_1$ Weight Normalization and 1-Path-Norm Regularization}




\begin{icmlauthorlist}
	\icmlauthor{Aditya Biswas}{}
\end{icmlauthorlist}


\icmlkeywords{Neural Networks, Path Norm, Weight Normalization, Capacity Control, Small Data, Weight Pruning, Sparsity}

\vskip 0.3in
]


\begin{abstract}
We present PSiLON Net, an MLP architecture that uses $L_1$ weight normalization for each weight vector and shares the length parameter across the layer.  The 1-path-norm provides a bound for the Lipschitz constant of a neural network and reflects on its generalizability, and we show how PSiLON Net's design drastically simplifies the 1-path-norm, while providing an inductive bias towards efficient learning and near-sparse parameters.  We propose a pruning method to achieve exact sparsity in the final stages of training, if desired.  To exploit the inductive bias of residual networks, we present a simplified residual block, leveraging concatenated ReLU activations. For networks constructed with such blocks, we prove that considering only a subset of possible paths in the 1-path-norm is sufficient to bound the Lipschitz constant.  Using the 1-path-norm and this improved bound as regularizers, we conduct experiments in the small data regime using overparameterized PSiLON Nets and PSiLON ResNets, demonstrating reliable optimization and strong performance.
\end{abstract}

\section{Introduction}

In direct contrast to classical statistical learning theory, the overparameterization of neural networks has been a driving factor towards its practical successes, even when using minimal explicit regularization. This suggests that functional capacity, or statistical complexity, is being controlled through other means as well, often dubbed \emph{inductive biases} \cite{inductive-bias}.  Techniques in this camp may not (or minimally) change the actual expressiveness of the function class; rather, they often modify the framing or dynamics of the optimization problem in some way. The breakthrough ResNet architecture \cite{resnet} was a prime example, combining several such inductive biases: identity skip connections, ReLU activations \cite{relu}, and batch normalization (BN) \cite{bn}.  

Understanding the mechanisms behind inductive biases and their relationship to network capacity is a crucial endeavor towards improving deep learning techniques and generalizability.  As a leading example, normalization layers have undergone many iterations of interpretation.  Weight normalization ($L_2$ WN) was introduced to explain the success of BN through a decoupling of learning length and direction \cite{wn}.  In fact, BN was shown to be nearly equivalent to $L_2$ WN, with BN normalizing the weights with a data-adaptive elliptical norm and both methods smoothing the loss landscape \cite{bn-expo-converge, bn-helps-opt}.  Empirically, it was also verified that BN allowed for much higher learning rates, and the same was theoretically proved for $L_2$ WN \cite{understanding-bn, normlayer-dynamics}. Finally, the persistent success of normalization layers pushed the community towards developing a theoretical understanding of their relation to functional capacity.  For example, $L_2$ WN was discovered to perform minimum-norm capacity control in overparameterized linear models \cite{wn-l2-minimum-norm}, and WN in deep neural networks was shown to control Rademacher complexity \cite{understanding-wn-dnn}.

Despite these profound discoveries, practical needs often demand a relationship in the opposite direction.  We would ideally want to \emph{start} with an understanding of network capacity that translates into an explicit equation or at least a reasonable bound. This could allow us to employ functional capacity control using hard constraints or a regularization term.  Only then would we want to choose useful, established inductive biases that remain effective and even \emph{synergize} with our strategy to ensure we produce controllable, generalizable networks that optimize with ease.

\paragraph{Contributions} 
Toward this end, we explore the interplay between certain architectural choices and the 1-path-norm \cite{capacity}. The 1-path-norm is an ideal candidate for a regularization term, as it bounds the Lipschitz constant of a neural network, which is a useful and practical measure of functional capacity.  Furthermore, path norms have been empirically demonstrated to correlate well with generalization, correctly capturing the effect of network depth \cite{fantastic}. Our primary aim is to design a network that complements the 1-path-norm.
\begin{itemize}
	\item We present $L_1$ Weight Normalization and motivate its strong inductive bias towards near-sparse vectors, for which we provide a definition as a continuous extension of bit sparsity.  We propose a method to achieve exact sparsity in a final pruning stage using $L_1$ Orthogonal Projection Weight Reparameterization.
	\item We design a simplified residual block consisting of only a concatenated ReLU activation \cite{crelu}, followed by a linear map. For residual networks composed of such blocks, we improve upon the bound given by the 1-path-norm by exploiting complementary sparsity patterns and only considering a subset of paths through the network. We also show that incorporating bias parameters is trivial.
	\item We show how $L_1$ WN with length parameter sharing across the layer greatly simplifies the 1-path-norm for multi-layer perceptron (MLP) networks and its improved bound for our MLP residual networks.  This makes the 1-path-norm cheap to compute and easy to optimize as a regularization term.  We provide empirical evidence that 1-path-norm regularization synergizes with WN to quickly discover subnetworks that generalize well.  We name our architectures \textbf{P}arameter \textbf{S}haring \textbf{i}n \textbf{L}-\textbf{O}ne \textbf{N}ormalized (PSiLON) Net and PSiLON ResNet.
	\item In the small data regime, we provide empirical evidence that overparameterized PSiLON Net outperforms a standard network architecture and, occasionally, random forests across a large suite of tabular datasets. We perform an ablation study using deep, overparameterized PSiLON ResNets on the FashionMNIST dataset and demonstrate the superior generalizability of models regularized with our improved bound as well as lower computational burden for our proposed model.
\end{itemize} 

\paragraph{Notation}
Vectors and matrices are denoted by lowercase and uppercase bold script, respectively. Neural networks will be denoted as $f_\mathbf{W}: \mathbb{R}^{d_\text{in}} \rightarrow \mathbb{R}^{d_\text{out}}$, where we use unsubscripted $\mathbf{W}$ to refer to the collection of all its parameters. Parameters corresponding to layer $k$ are subscripted as $\mathbf{W}_k$, and hidden layer activations are denoted with superscripts $\mathbf{z}^{(k)}$.  Scalar functions applied to vectors or matrices are understood to be applied elementwise.  We denote the (sub-)gradient and Jacobian with respect to $\mathbf{v}$ both by $\nabla_\mathbf{v}$.  We let $\lVert \cdot \rVert_p$ represent both the vector $L_p$ norm and matrix operator norm induced by the vector $L_p$ norm. $\mathbf{I}, \mathbf{1},$ and $\mathbf{0}$ will refer to the identity matrix, one vector, and zero vector, respectively.  We denote their dimension by a subscript unless it can be clearly inferred from context.


\section{$L_1$ Weight Normalization}
For clarity of exposition, consider the basic linear model under a similar reparameterization to the original $L_2$ weight normalization (WN) technique \cite{wn}:
\begin{align}
f(\mathbf{x}) &= \mathbf{w}^T \mathbf{x} \quad \text{where} \quad \mathbf{w} = \frac{g}{\lVert \mathbf{v} \rVert_1} \mathbf{v}.
\intertext{Like $L_2$ weight normalization, we take $g \in \mathbb{R}$ rather than enforcing positivity. The subgradient with respect to $\mathbf{v}$ is}
\nabla_\mathbf{v} f(\mathbf{x}) &= \frac{g}{\lVert \mathbf{v} \rVert_1} \bigg(\mathbf{x} - \frac{ \mathbf{w}^T \mathbf{x}}{\lVert \mathbf{v} \rVert_1} \text{sign}(\mathbf{v}) \bigg) \nonumber\\
&= \frac{g}{\lVert \mathbf{v} \rVert_1} M_\mathbf{w} \mathbf{x} \\
\text{where} \quad  M_\mathbf{w} &= \mathbf{I} - \tfrac{1}{\lVert \mathbf{w} \rVert_1} \text{sign}(\mathbf{w})\mathbf{w}^T. \nonumber
\end{align}
$M_\mathbf{w}$ acts as a projection matrix onto $\mathbf{w}^\perp$, the orthogonal complement of $\text{span}\{\mathbf{w}\}$; idempotence and the stated orthogonality can easily be verified after noticing $\lVert \mathbf{w} \rVert_1 = \mathbf{w}^T \text{sign}(\mathbf{w})$.  The gradient using $L_2$ WN has a nearly identical form, simply substituting with the $L_2$ norm and using an orthogonal projection onto $\mathbf{w}^\perp$.  In contrast, $M_\mathbf{w}$'s projection is oblique.  Specifically, when applied on $\mathbf{x}$, it projects along the line $\mathbf{x}+t \text{sign}(\mathbf{w})$ with free parameter $t$. Even so, since $\nabla_\mathbf{v} f(x) \perp \mathbf{w}$, we inherit $L_2$ WN's properties of increasing but self-stabilizing norm and associated robustness to the learning rate when using gradient-based optimization. 

\subsection{Inductive Bias}
Since our oblique projection is discontinuous in $\mathbf{w}$, optimization dynamics will behave uniquely in comparison to $L_2$ WN.  Let us write this oblique projection of $\mathbf{x}$ coordinate-wise when limits are taken towards 0:
\begin{align*}
	 \lim_{w_i \rightarrow 0^-} \big(M_\mathbf{w} \mathbf{x} \big)_i  &=  x_i + \frac{\mathbf{w}^T \mathbf{x}}{\lVert \mathbf{w} \rVert_1} \\
	 \lim_{w_i \rightarrow 0^+} \big(M_\mathbf{w} \mathbf{x} \big)_i &= x_i - \frac{\mathbf{w}^T \mathbf{x}}{\lVert \mathbf{w} \rVert_1}.
\end{align*}
From this, we can see that as $w_i$ approaches and crosses the boundary of 0, $\tfrac{\partial}{\partial v_i} f(x)$ may switch signs if $|x_i| < \lvert \mathbf{w}^T \mathbf{x} \rvert / \lVert \mathbf{w} \rVert_1$, depending on the signs of $\mathbf{w}^T \mathbf{x}$ and $x_i$, leading to 0 being an attracting point.

When applying our reparameterization to deep neural networks, this creates a further inductive bias for ``near-sparse" learning. Even if there is no sign change, a parameter may approach 0 quickly, but after crossing its trajectory immediately slows and allows the rest of the network an opportunity to adapt around its more stable value.  Alternatively, a parameter may approach 0 slowly, giving the network a chance to prefer its value near 0, but cross and rapidly move away if the network fails to do so.  

When optimization steps use mini-batch gradient descent, the described cases will align with different samples in the batch. Therefore, batch size and learning rate may play non-negligible roles.  Also, one may reasonably fear that large, discrete changes in the gradient across axes could destabilize learning.  Gradient clipping could provide a solution for this, but we found it to be unnecessary in early experiments.

In summary, while $L_1$ WN inherits the effective learning rate arguments of $L_2$ WN, it additionally changes the dynamics of optimization such that the network spends more time exploring near-sparse settings, which we shall now define. Exact sparsity for bit vectors $\mathbf{v} \in \{0,1\}^d$ can be defined as $\text{Sparsity}(\mathbf{v}) = 1 - \tfrac{1}{d} \mathbf{1}^T\mathbf{v}$. We define near-sparsity to be a continuous extension of exact sparsity for bit vectors. \\
\begin{definition}[Near-Sparsity]
For a weight vector $\mathbf{v} \in \mathbb{R}^d$, construct a probability vector representing a Categorical distribution as follows: $p(\mathbf{v}) = \lvert \mathbf{v} \rvert / \lVert \mathbf{v} \rVert_1$. Let $\mathbb{H}[p(\mathbf{v})]$ be the distribution's entropy using the natural log.  Then we say 
\begin{equation}
	\text{NSparsity}(\mathbf{v}) = \begin{cases} 1 & \text{if} \;\; \mathbf{v} = \mathbf{0} \\
		1- \frac{1}{d} \exp \{\mathbb{H}[p(\mathbf{v})] \} & \text{otherwise}.
	\end{cases} \label{sparsitydef}
\end{equation}
For the entropy calculation, we define $0 \ln (0) = 0$ by right continuity to allow for zero entries.
\end{definition}
\subsection{Weight Pruning} \label{pruning method}
Exact sparsity in the network weights may be desired for reducing memory or compute in downstream applications.  To achieve this goal we propose a reparameterization that achieves exact sparsity: $L_1$ Orthogonal Projection Weight Reparameterization.  Let $\text{Proj}_{L_1}$ represent the orthogonal projection operator onto the $L_1$ unit sphere. We reparameterize the weight vector as
\begin{equation}
\mathbf{w} = g \text{Proj}_{L_1}( \mathbf{v}).
\end{equation}
A subdifferentiable algorithm for this projection is provided in Appendix \ref{L1 proj appendix}.  One may consider using this reparameterization over $L_1$ WN from the start of training.  However, a major drawback of this approach is that the gradients are sparse for large regions of the parameter space, which can impede learning since parameters can stop receiving a gradient signal and become trapped.  We suggest first learning with $L_1$ WN to explore the parameter space more thoroughly.

Near the end of training, we propose initiating a pruning stage where we form a convex combination of the two projections using $\alpha \in [0,1]$:
\begin{equation}
	\mathbf{w} =  (1-\alpha) \frac{g}{\lVert \mathbf{v} \rVert_1} \mathbf{v} + \alpha g  \text{Proj}_{L_1}( \mathbf{v}).
\end{equation}
We linearly anneal $\alpha$ from 0 to 1 over the pruning stage such that the convex combination gradually switches from using an oblique projection to an orthogonal projection by the end of training, producing exact sparsity.


\section{1-Path-Norm}
Consider a neural network consisting of $K$ weight matrices $\mathbf{W}_k$, interleaved with activation functions $\sigma$ and final output nonlinearity $\sigma_\text{out}$:
\begin{align}
f_\mathbf{W}(\mathbf{x}) &= \sigma_\text{out} (\mathbf{W}_K \sigma(\mathbf{W}_{K-1} \sigma(\ldots\sigma(\mathbf{W}_1 \mathbf{x}) \ldots))). \label{nn eq}
\intertext{This structure is also known as a multi-layer perceptron (MLP).  The 1-path-norm can be computed as}
P_1(\mathbf{W}) &= \mathbf{1}^T \lvert \mathbf{W}_K \rvert \lvert \mathbf{W}_{K-1} \rvert \ldots  \lvert \mathbf{W}_1 \rvert \mathbf{1}. \label{1-path-norm}
\end{align}
and is equivalent to its namesake definition of collecting every unique path on the graph from input to output, taking the product of weights along each path, and finally summing the absolute values of the results \cite{capacity}.  \citet{1pathnorm-dnn} gives us the following theorem (with minor modification for multiple outputs and final nonlinearity):
\begin{theorem}
	\label{lemma 3}
	 Suppose the subgradients of $\sigma$ and $\sigma_\text{out}$ are globally bounded between zero and one.  Let $\mathcal{L}_\mathbf{W}$ denote the Lipschitz constant of the neural network $f_\mathbf{W}$ with respect to the $L_\infty$ and $L_1$ norms for the input and output spaces, respectively. Then,
	\begin{equation}
	\mathcal{L}_\mathbf{W}  \leq P_1(\mathbf{W}) \leq \lVert \mathbf{W}_K \rVert_{\infty,1} \prod_{k=1}^{K-1} \lVert \mathbf{W}_k \rVert_\infty 
	\end{equation}
\end{theorem}

This theorem provides us a means for achieving control over the network's Lipschitz constant. As the bound given by the 1-path-norm is subdifferentiable, it would make for an ideal regularization candidate. We are furthermore guaranteed that this bound is at least as good as the trivial operator product bound and, in practice, often substantially smaller.

\subsection{Residual CReLU Networks}
Modern neural networks employ many additional techniques to create inductive biases that improve optimization and lead toward generalization. The first technique we incorporate is residualization, i.e. the usage of identity skip connections.  For our residual networks, we assume all hidden layers have dimension $d$.  

Using the $k$-th hidden layer's representation $\mathbf{z}^{(k)}$ as input, residual blocks are typically implemented as follows:
\begin{equation}
\mathbf{z}^{(k+1)} = \mathbf{z}^{(k)} + \mathbf{W}_{k,2} \sigma(\mathbf{W}_{k,1} \mathbf{z}^{(k)}). \nonumber
\end{equation}
Instead, we propose to use the concatenated ReLU (CReLU) activation function \cite{crelu} to simplify a block back to the standard pattern of a nonlinearity followed by a linear map. 

\paragraph{CReLU Residual Block} Let $\sigma^+(\mathbf{z}) = \text{ReLU}(\mathbf{z})$ and $\sigma^-(\mathbf{z}) = -\text{ReLU}(-\mathbf{z})$.  We define CReLU to apply the following map: 
\begin{equation*}
\sigma: \mathbf{z} \rightarrow \begin{bmatrix} \sigma^+(\mathbf{z}) \\ \sigma^-(\mathbf{z}) \end{bmatrix}
\end{equation*}
As an aside, this activation has the non-negative homogeneous property $\sigma(c\mathbf{z}) = c\sigma(\mathbf{z})$ for $c \geq 0$, which allows us to use similar rescaling arguments for ReLU networks that were first used to motivate the 1-path-norm. CReLU additionally solves an issue with ReLU activations in that they fail to propagate gradient information for negative inputs and can ``die" if all inputs become negative.  A drawback of CReLU is that it doubles the size of each weight matrix. Let $\mathbf{W}^+, \mathbf{W}^-$ be the weights associated with the $\sigma^+, \sigma^-$ activations, respectively.  We propose the following update equation for layer $k+1$:
\begin{align}
\mathbf{W}_k &=  \begin{bmatrix} (\mathbf{I}+\mathbf{W}^+_k) & (\mathbf{I} + \mathbf{W}^-_k) \end{bmatrix} \nonumber \\
\mathbf{z}^{(k+1)} &=  \mathbf{W}_k  \sigma\big(\mathbf{z}^{(k)} \big)  \label{skip_mat}\\
&= \mathbf{z}^{(k)} + \mathbf{W}^+_k \sigma^+(\mathbf{z}^{(k)}) + \mathbf{W}^-_k \sigma^-(\mathbf{z}^{(k)}). \label{skip}
\end{align}
Equation \ref{skip_mat} shows us that we can still express the 1-path-norm as simply as before in theory, while Equation \ref{skip} shows that this structure gives us our desired identity skip connection.  However, the number of paths through the network now doubles with each consecutive layer, leading to a computational complexity of $\mathcal{O}(2^K K d^2)$ for computing the 1-path-norm.  For comparison, the 1-path-norm of an MLP scales as $\mathcal{O}(K d^2)$, which is linear in depth.
\begin{definition}[CReLU Residual Network]
A CReLU Residual Network is a neural network $f_\mathbf{W}$ composed of an initial linear map, followed by consecutive CReLU residual blocks as given in Equation \ref{skip_mat}, followed by a CReLU activation and a final linear map, and ending with a nonlinearity $\sigma_\text{out}$.
\end{definition}

\paragraph{Improved Bound} CReLU produces a complementary sparsity pattern, which we exploit to produce a better bound for the Lipschitz constant than a naive application of Equation \ref{1-path-norm}. Furthermore, this allows us to restore the computational complexity to once again being linear with depth. For a CReLU residual network, let $\mathbf{\tilde{W}}_k = \text{max}(\lvert \mathbf{W}^+_k \rvert, \lvert \mathbf{W}^-_k \rvert)$ and define
\begin{equation}
	\tilde{P}_1(\mathbf{W}) =  \mathbf{1}^T \mathbf{\tilde{W}}_K \big( \mathbf{I} + \mathbf{\tilde{W}}_{K-1}  \big) \ldots \big( \mathbf{I} + \mathbf{\tilde{W}}_2 \big) \lvert \mathbf{W}_1 \rvert \mathbf{1} \label{1pathnorm-crelu}
\end{equation}
\begin{theorem}
	 For a CReLU residual network $f_\mathbf{W}$, suppose $P_1(\mathbf{W})$ computes the 1-path-norm where skip paths and weight paths are treated as distinct. Suppose the subgradient of $\sigma_\text{out}$ is globally bounded between zero and one.  Let $\mathcal{L}_\mathbf{W}$ denote the Lipschitz constant of $f_\mathbf{W}$ with respect to the $L_\infty$ and $L_1$ norms for the input and output spaces, respectively. Then,
	\begin{equation}
		\mathcal{L}_\mathbf{W} \leq \tilde{P}_1(\mathbf{W}) \leq P_1(\mathbf{W}).
	\end{equation}
\end{theorem}
with a proof given in Appendix \ref{proof appendix}.  $\tilde{P}_1(\mathbf{W})$ is likely to produce a tighter bound in practice with a CReLU residual network than $P_1(\mathbf{W})$ is to give for an equivalently sized ReLU residual network.


\subsection{Bias Parameters}
A less impactful but prevalent architectural feature is the presence of bias or shift parameters, which we incorporate next.  The 1-path-norm definition can easily be adjusted to account for bias parameter vectors $\mathbf{b}_k$ by a simple extension of Equation \ref{1-path-norm}:
\begin{align}
	P_1(\mathbf{W}) &=  \mathbf{1}^T_{d_\text{out}} \lvert \mathbf{\hat{W}}_K \rvert \lvert \mathbf{\hat{W}}_{K-1} \rvert\ldots  \lvert \mathbf{\hat{W}}_1 \rvert \begin{bmatrix} 0 & \mathbf{1}_{d_\text{in}}^T \end{bmatrix}^T \label{bias} \\
	\text{where} & \quad \mathbf{\hat{W}}_k = \begin{bmatrix}  1 & \mathbf{0}^T \\ \mathbf{b}_k  & \mathbf{W}_k  \end{bmatrix} \quad \text{for} \; k \in \{1,\ldots,K-1\} \nonumber \\
	\text{and} & \quad \mathbf{\hat{W}}_K =\begin{bmatrix}  \mathbf{b}_K  & \mathbf{W}_K  \end{bmatrix}. \nonumber
\end{align}
This is a simple consequence of bias parameters corresponding to an extra neuron at each layer; each such neuron receives a single connection with weight 1 from the previous layer's bias neuron. The input bias neuron cannot vary, resulting in the 0 scalar in Equation \ref{bias}.  This structure shows that we can attain the same result by simply ignoring bias parameters and directly applying Equation \ref{1-path-norm}. This would also clearly work with Equation \ref{1pathnorm-crelu}.

\subsection{Regularization} \label{reg section}
The final and most important inductive bias technique we incorporate is the use of normalization layers. Our method will not only improve optimization, but crucially provide a manageable regularization term.  We begin by describing our method with respect to multi-layer perceptrons (MLPs) before adapting the method to work for CReLU residual networks.

\paragraph{MLPs}
We propose to use $L_1$ WN for each row of $\mathbf{W}_k \; \forall k$.  Collect the $g$ length parameters associated with each row of $\mathbf{W}_k$ into a vector  $\mathbf{g}_k$, and let the row-normalized matrix be denoted $\mathbf{V}_k$, such that $\mathbf{W}_k = \text{diag}(\mathbf{g_k}) \mathbf{V}_k $.  Under this reparameterization, Equation \ref{1-path-norm} for the 1-path-norm can be expressed as
\begin{align}
	P_1(\mathbf{W}) &= \lvert \mathbf{g}^T_K \rvert \lvert \mathbf{V}_K \rvert  \ldots \lvert \text{diag}(\mathbf{g}_1) \rvert \lvert \mathbf{V}_1 \rvert  \mathbf{1}.
	\intertext{For layers $k \in \{1,\ldots, K-1\}$, if we enforce parameter sharing for the length parameter across a layer, i.e. $\mathbf{g}_k = g_k \mathbf{1}$ for $g_k \in \mathbb{R}$, this simplifies to}
	P_1(\mathbf{W}) &=  \lVert \mathbf{g}_K \rVert_1  \prod_{k=1}^{K-1} \lvert g_k  \rvert. \label{l1 path norm}
\end{align}
We name this network design \textbf{P}arameter \textbf{S}haring \textbf{i}n \textbf{L}-\textbf{O}ne \textbf{N}ormalized (PSiLON) Net.

This design could be simplified even further by freezing $g_k = 1$ for $k \in \{1,\ldots K-1\}$. Indeed, such an approach is suggested in \citet{understanding-wn-dnn}.  However, when $g_k=1$, its associated linear layer may shrink the norm of its incoming layer due to the nonlinearity; this effect can grow exponentially with depth early in training, slowing convergence. Thus, such a choice would only be acceptable for shallow networks. Allowing only $g_1$ to be free could partially compensate for this negative effect.  We do not pursue this approach in this work.

\paragraph{CReLU Residual Networks}
This strategy is straightforward to adapt for our residual networks.  We start with Equation \ref{1pathnorm-crelu}.  Then to simplify, $L_1$ WN with parameter sharing proceeds as above for $\mathbf{W}_1$.  For all other layers, we normalize the rows of both $\mathbf{W}^+_k$ and $\mathbf{W}^-_k$ by the $L_1$ norm of the rows of $\mathbf{\tilde{W}}_k$.  A common scalar length parameter $g_k$ is shared between both matrices, except at the final layer where a common vector parameter $\mathbf{g}_K$ is shared.  Incorporating this network design, we achieve a similar simplification to earlier:
\begin{equation}
	\tilde{P}_1(\mathbf{W}) = \lVert \mathbf{g}_K \rVert_1 \lvert g_1 \rvert \prod_{k=2}^{K-1} (1+ \lvert g_k \rvert). \label{l1 path norm res}
\end{equation}
We name this network design PSiLON ResNet.

\paragraph{Remarks}
Regularizing the loss function with this expression, we can maintain control over the Lipschitz constant of the network, while also alleviating the potential optimization difficulty that a direct implementation of Equations \ref{1-path-norm} or \ref{1pathnorm-crelu} as regularizers may have.  This also greatly relieves the computational burden of repeatedly computing the 1-path-norm or the improved bound at every optimization step.  In fact, we have improved the computational complexity to a mere $\mathcal{O}(K)$.  

Under this design, the 1-path-norm becomes equivalent to the trivial product bound for the Lipschitz constant using the operator $\infty$-norm.  However, this should not concern us: this is the best possible case for the product bound. Because the product bound is exactly the 1-path-norm here, the interactions of weights between different layers is still captured to produce a meaningful bound.

For PSiLON ResNet, note that this design leads to the rows of the normalized but unscaled matrices $\mathbf{V}_k^+$ and $\mathbf{V}_k^-$ now representing points within the $L_1$ ball rather than exactly on the sphere.


\section{Experiments}

\subsection{Small Tabular Datasets} \label{tabular experiment}
\paragraph{Data Selection and Preprocessing}
We borrow the tabular dataset suite curated in \citet{datasets} for both regression and classification. This collection only contains independently and identically distributed data and sufficiently preprocesses much of the data for us.  Additionally, skewed regression targets are log-transformed and multi-classification tasks are coarsened to binary tasks. We only keep datasets with more than 10 predictors and exclude classification datasets that also appear as a regression task, leaving 17 regression and 13 classification datasets. For larger datasets, we randomly subset on 20,000 samples.  To explore the benefits of capacity control, we restrict our training set to 2,000 samples per dataset.  All remaining data is split evenly between validation and test sets.  Numerical predictors are standardized by their mean and standard deviation, and regression targets are standardized by their median and quartile deviation to normalize the reasonable range of regularization parameters across datasets.

\paragraph{Model Specifications}
Our primary model is PSiLON Net (P-Net) using 1-path-norm regularization in the form of Equation \ref{l1 path norm}.  We compare this to an MLP (S-Net) using $L_2$ weight normalization (WN) and $L_2$ weight regularization (WR).  As $L_2$ WN is theoretically close to batch normalization and $L_2$ WR is nearly identical to weight decay, these choices align closely to a ``standard" network a practitioner may employ.  Note that S-Net has slightly more representational power than P-Net, since it does not enforce any parameter sharing.  For both networks, we overparameterize with 3 hidden layers of 500 neurons each so that the model has slightly over 0.5M parameters excluding the first linear map; this necessitates the need for capacity control.  We do not use early stopping as this would obfuscate the role of capacity control.  Instead, we fix a compute budget of 5,000 optimization steps and perform a grid search for the optimal regularization hyperparameter using the validation set.  Additional training details are provided in Appendix \ref{small tabular appendix}.  

\paragraph{Results}
Root mean squared error (RMSE) and cross-entropy loss for the test set in regression and classification tasks are provided in Tables \ref{regression} and \ref{classification}, respectively, where we shorten some dataset names and report the better performance between the two models in bold.  We additionally provide results of hyperparameter-tuned $L_2$ regularized linear/logistic regression (LR) and random forests (RF) as baselines for interpretation.

The results for both networks are surprisingly often comparable with random forests, a gold standard model for tabular data. In fact, P-Net beats the RF baseline in 15/30 datasets and maintains close performance in 10/15 of its losses. This stands in contrast to much of the literature and popular wisdom; this discrepancy could be partially explained through an insufficient degree of overparameterization and hyperparameter search in other settings.  Between the two neural networks, P-Net has better performance than S-Net in 21/30 datasets with an improvement greater than $0.01$ in 15 of them. S-Net has an improvement greater than $0.01$ in 3 datasets.  This suggests that networks with their capacity restricted via the 1-path-norm more closely align with function classes present in real-world tabular data than networks with $L_2$ weight regularization.  This also verifies that such networks are effectively trainable.

\begin{table}
\caption{Test set RMSE for regression datasets.}
\label{regression}
\vskip 0.1in
\begin{center}
\begin{small}
\begin{sc}
\begin{tabular}{|c || c c | c c |} 
 \hline
 Dataset & LR & RF & S-Net  & P-Net \\ 
 \hline\hline
 cpu\_act & 1.460	& 0.428	& 0.403 & \textbf{0.385} \\ 
 \hline
 pol & 0.878 & 0.243 & \textbf{0.111} & 0.147 \\
 \hline
 elevators & 1.202 & 1.382 & 0.793 & \textbf{0.787} \\
 \hline
 wine\_quality & 1.509 & 1.384 & 1.449 & \textbf{1.435} \\
 \hline
 Ailerons & 0.707 & 0.723 & 0.665 & \textbf{0.662} \\
 \hline  
 houses\_16H & 1.735 & 1.401 & 1.477 & \textbf{1.459} \\
 \hline
 MiamiHousing & 0.841 & 0.566 & 0.518 & \textbf{0.509} \\
 \hline
 superconduct & 0.721 & 0.564 & \textbf{0.579} & 0.598 \\
 \hline
 yprop & 2.723 & 2.715 & 2.754 & \textbf{2.705} \\
 \hline 
 benz\_manufact & 1.015 & 0.998 & 1.004 & \textbf{0.992} \\
 \hline   
 brazillian\_house & 0.518 & 0.100 & 0.273 & \textbf{0.136} \\
 \hline
 bike-sharing & 1.194 & 0.517 & 0.474 & \textbf{0.456} \\
 \hline
 nyc-taxi & 1.477 & 1.257 & \textbf{1.266} & 1.316 \\
 \hline  
 house\_sales & 0.726 & 0.600 & 0.568 & \textbf{0.541} \\
 \hline    
 topo & 2.772 & 2.703 & 2.848 & \textbf{2.720} \\
 \hline   
 delays\_transport & 0.981 & 0.975 & \textbf{0.974} & 0.975 \\
 \hline   
 allstate\_claims & 1.014 & 1.030 & \textbf{1.019} & 1.021 \\
 \hline   
\end{tabular}
\end{sc}
\end{small}
\end{center}
\vskip -0.1in
\end{table}

\begin{table}
	\caption{Test set cross-entropy loss for classification datasets.}
	\label{classification}
	\vskip 0.1in
	\begin{center}
		\begin{small}
			\begin{sc}
				\begin{tabular}{|c || c c | c c |} 
					\hline
					Dataset & LR & RF & S-Net  & P-Net \\ 
					\hline\hline
					credit & 0.585 & 0.493 & 0.529 & \textbf{0.526} \\ 
					\hline
					MagicTelescope & 0.4938 & 0.376 &	\textbf{0.357} & 0.358 \\
					\hline
					MiniBooNE & 0.294 & 0.255 & \textbf{0.256} & 0.261 \\
					\hline
					Higgs & 0.645 & 0.602 & 0.693 & \textbf{0.616} \\
					\hline
					jannis & 0.558 & 0.525 & 0.538 & \textbf{0.526} \\
					\hline  
					Bioresponse & 0.556 & 0.505 & 0.557 & \textbf{0.521} \\
					\hline
					heloc & 0.578 & 0.557 & 0.567 & \textbf{0.564} \\
					\hline
					eye\_movements & 0.681 & 0.662 & 0.694 & \textbf{0.670} \\
					\hline
					covertype & 0.486 & 0.475 & \textbf{0.456} & 0.458 \\
					\hline 
					albert & 0.650 & 0.628 & 0.653 & \textbf{0.640} \\
					\hline   
					default\_credit & 0.612 & 0.565 & 0.571 & \textbf{0.569} \\
					\hline
					road\_safety & 0.590 & 0.530 & 0.613 & \textbf{0.520} \\
					\hline
					compas & 0.606 & 0.603 & \textbf{0.594} & 0.594 \\
					\hline   
				\end{tabular}
			\end{sc}
		\end{small}
	\end{center}
	\vskip -0.1in
\end{table}

\begin{figure}[ht]
	\vskip 0.1in
	\begin{center}
		\centerline{\includegraphics[width=\columnwidth]{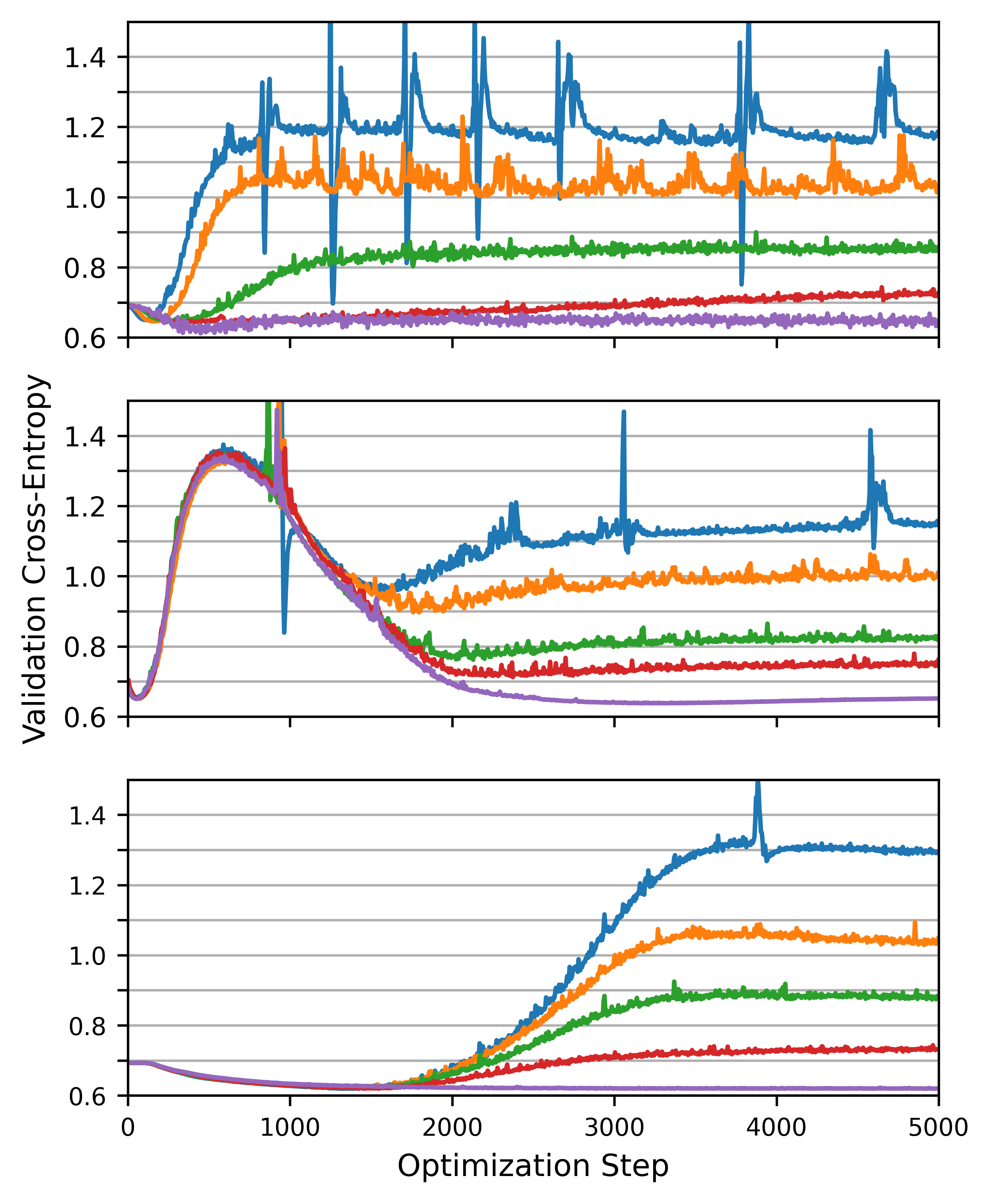}}
		\caption{Validation cross-entropy curves on the Higgs dataset under varying regularization levels indicated by color. Using $L_2$ WR, results are presented for S-Net without WN (top) and S-Net with WN (middle),  Using 1-path-norm regularization, results are presented for P-Net (bottom).}
		\label{curves}
	\end{center}
	\vskip -0.1in
\end{figure}

\subsection{Higgs Dataset Case Study} 
\paragraph{Generalization across Training Period}
We cherry-pick a dataset where P-Net performed significantly better than S-Net to explore the dynamics of model fitting and effect of capacity control in explaining the discrepancy.

Figure \ref{curves} shows the validation cross-entropy across the training run for each network across 5 hyperparameter values; an $L_2$ WR network without WN is also provided for comparison. A more fine-grained hyperparameter search was performed for the $L_2$ WR networks to see if this could account for the gap.  While we were able to improve test set performance to $0.646$ in both $L_2$ WR networks (shown in purple), hyperparameter choice alone could not fully explain the discrepancy.  As this value was achievable by both $L_2$ WR models, this also suggests that the method of capacity control may play a more important role here than the optimization dynamics. Different network initializations failed to make a meaningful difference in these results.

Both $L_1$ and $L_2$ WN stabilize the dynamics of the function over the training period as expected.  However, S-Net presents with a unique learning pattern, where it very quickly and excessively overfits the data and then employs its capacity control to smooth the function afterwards. On the other hand, P-Net exhibits ``textbook" learning dynamics, where there is an underfitting stage smoothly transitioning to an overfitting stage.  Furthermore, all P-Net models converge to nearly the optimal value achievable at the transition point. We observed similar behavior for most datasets considered in this study.  Such learning dynamics inspire confidence in the practical use of P-Net, and we leave it to future work to explore the theoretical reasons behind this behavior in contrast to S-Net's dynamics.

\paragraph{Weight Pruning}
We use Definition 1 for near-sparsity to explore P-Net's potential for weight pruning.  We call a network's near-sparsity the average near-sparsity across all weight vectors. We continue to use the Higgs dataset for this experiment.  For the optimal models, S-Net has an NSparsity of $0.169$, and P-Net has an NSparsity of $0.725$. Furthermore, P-Net very quickly moves towards this level of NSparsity early in training. This difference cannot be completely attributed to different regularization strategies.  An unregularized P-Net has an NSparsity of $0.496$ for this dataset, supporting the hypothesis that $L_1$ WN does indeed have its own inductive bias towards near-sparse weight vectors in neural networks with 1-path-norm regularization amplifying this effect.  Keeping the total number of update steps fixed, we modify the batch size by $0.25 \times$ and $5\times$, which produces an NSparsity of $0.609$ and $0.749$, respectively, supporting the hypothesis that batch size has a weak but noticeable effect on the near-sparsity level.

As P-Net has a high near-sparsity, we use the techniques described in Section \ref{pruning method} to prune the network.  We train the network as before for 4,000 parameter updates, but transition to the pruning strategy for the final 1,000 updates.  After pruning is complete, the final network has an exact sparsity of $0.817$ and a slightly improved test set cross-entropy of $0.614$, meaning we could prune over 0.42M parameters from this model with no loss of performance.  In the language of the lottery ticket hypothesis \cite{lottery}, these results demonstrate that P-Net automatically and quickly converges towards a ``winning ticket" (i.e. a subnetwork present at initialization that would generalize well in isolation) and simply requires a final pruning stage to technically meet the criteria of exactly sparsity.  Using this pruning strategy from the beginning of training produces notably less sparse and performant networks as hypothesized.

\begin{table}
	\caption{Training cross-entropy (CE-Tr), test cross-entropy (CE-Te), test accuracy (Acc-Te), final near sparsity (NS) and relative to P-ResNet wall-clock training time (Time) for each model. ``ResNet" is suppressed from each model name.}
	\label{deep table}
	\vskip 0.15in
	\begin{center}
		\begin{small}
			\begin{sc}
				\begin{tabular}{|c || c |  c c | c c |} 
					\hline
					Model & CE-Tr & CE-Te & Acc-Te  & NS & Time \\ 
					\hline\hline
					P & 0.098 & 0.408 & 0.859  & 0.892 & $1.00 \times$ \\ 
					\hline
					L1-IB & 0.112 & 0.419 & 0.859 & 0.802 & $1.25 \times$ \\ 
					\hline
					L2-IB & 0.083 & \textbf{0.403} & \textbf{0.864}  & 0.861 & $1.25 \times$ \\ 
					\hline
					L1-OB & 0.314 & 0.477 & 0.841 &   0.596 & $1.35 \times$ \\ 
					\hline
					L2-OB & 0.217 & 0.439 & 0.859 &   0.874 & $ 1.35 \times$ \\ 
					\hline
					S & 0.315 & 0.492 & 0.835 & 0.693 & $1.10 \times$ \\ 
					\hline
					Pj &  0.298 & 0.420  & 0.851   & 0.811 & $1.00 \times$ \\ 
					\hline
				\end{tabular}
			\end{sc}
		\end{small}
	\end{center}
	\vskip -0.1in
\end{table}

\subsection{Ablation Study on Deep Networks} \label{deep experiment}
\paragraph{Data Selection and Preprocessing}
For our next experiment, we use the Fashion-MNIST dataset.  To necessitate the need for capacity control, we randomly subset the original training set to half its samples and allocate 10,000 samples to the training set and 20,000 to the validation set.  We use the original test set of 10,000 samples and the original learning objective of classification across 10 categories. We flatten the image across the height and width dimensions and rescale each feature to lie in the range $[-1,1]$.

\paragraph{Model Specifications}
We ablate or modify several choices in the PSiLON residual network (P-ResNet) architecture regularized with our improved bound in the form of Equation \ref{l1 path norm res}.  We consider CReLU residual networks with $L_1$ normalization but no parameter sharing, regularized by either the improved bound given in Equation \ref{1pathnorm-crelu} or the original 1-path-norm bound given by Equation \ref{1-path-norm}, referred to as L1-IB-ResNet and L1-OB-ResNet, respectively.  We consider similar networks except using $L_2$ normalization, referred to as L2-IB-ResNet and L2-OB-ResNet, respectively.  We include another ``Standard" Net with $L_2$ WN and $L_2$ WR, adapted for the CReLU residual architecture and referred to as S-ResNet.  Lastly, we implement a version of P-ResNet where the pre-length-scaled weight vectors are constrained within the $L_1$ ball, which we call Pj-ResNet. We train this model using a projected gradient descent version of our optimization strategy.  For all models, the normalization is done using the $L_1$ or $L_2$ norm of the rows of $\mathbf{\tilde{W}}_k$ as described in Section \ref{reg section}, regardless of the regularization strategy used.

For all networks, we overparameterize with 10 hidden layers of 500 neurons each, so that each model has approximately 5M parameters. We fix a compute budget of 20,000 optimization steps and perform a grid search for the optimal regularization hyperparameter using the validation set.  Additional training details are provided in Appendix \ref{ablation deep appendix}.

\paragraph{Results}
Table \ref{deep table} presents train and test set performance metrics of each model as well as their final near-sparsities.  The relative wall-clock time is also presented when training using a single NVIDIA GeForce GTX 1070 Ti GPU with PyTorch JIT-compiled networks \cite{pytorch}.  

Models trained with the improved bound regularization term had the best performances and were quite close to each other.  As a first negative result, we observe that gradient descent optimization using the full regularization term (L1-IB and L2-IB-ResNets) is not as difficult as previously assumed, at least under our settings.  As a second negative result, we observe that $L_2$ WN models marginally outperform and achieve higher near-sparsities than their $L_1$ WN counterparts.  However, this can be mostly explained by our observation that $L_1$ WN models experienced slightly slower rates of convergence than $L_2$ WN models, so the former end up being marginally undertrained in comparison given our fixed compute budget.  Pj-ResNet experienced difficulties with optimization, explaining its slightly lower final performance as well.  Excluding Pj-ResNet, it is notable that the models in this group were able to achieve very good fits to the training set, while still being able to generalize well.  This can be interpreted as evidence that we are not simply indiscriminately restricting the functional capacity but rather controlling it in a more data-adaptive manner.

As hypothesized, using the original 1-path-norm formulation (L1-OB and L2-OB-ResNets) results in noticeably worse performance as the bound is looser for this architecture.  S-ResNet presents with the weakest results, reaffirming that the 1-path-norm is a superior quantity for measuring network capacity.

The performance difference in comparison to L2-IB-ResNet is least pronounced in P-ResNet, which also enjoys a significantly faster wall-clock time due to its simple regularization term.  Due to these reasons, P-ResNet may achieve the best performance under a compute budget given in wall-clock time.

\section{Related Work}

Similar in spirit to our work is \citet{lipschitz-enforce}, who explore capacity control through projection-based weight normalization ($L_1$, $L_2$, and $L_\infty$) with allowances for modern features such as convolutions and residual layers.  However, their work differs in two crucial ways. First, they employ capacity control by constraining the looser operator product bound. Second, they report difficulties with optimization when penalizing via a regularization term, so they opt to enforce a hard constraint on the weight norms and use projected gradient descent. \citet{orthogonal-lipschitz-networks} also achieve capacity control through constraining the operator product bound, but use Almost-Orthogonal Lipschitz layers to mitigate some of its drawbacks, which are a special reparameterization technique to guarantee the layer is 1-Lipschitz with respect to the Euclidean norm. \citet{groupsort} seeks to learn 1-Lipschitz neural networks by proposing a gradient norm preserving activation function GroupSort and enforcing $L_2$ and $L_\infty$ norm-constrained linear maps. \citet{projection-wn} places an $L_2$ unit norm constraint on the network weights for the purpose of improving the conditioning of the Hessian; they operationalize this through a projected gradient descent method and report good results combining this with batch normalization to recover the model's capacity. Lastly, \citet{implicit-bias-geometry} studies capacity control through the implicit biases that emerge under different choices of step size and direction in iterative optimization.

Another major line of work incorporates path-norms as regularizers, often making use of specialized optimization techniques.  \citet{path-sgd} derives the steepest descent direction with respect to the path norm. \citet{conditional-gradient} improves upon this technique with a conditional gradient method, notably showing strong, stable performance regularizing the 2-path-norm in deep residual networks.  \citet{prox-shallow, 1pathnorm-dnn} optimize their networks using proximal algorithms.  They report good results when using 1-3 hidden layers in fully-connected networks, but note that the advantage over direct auto-differentiation methods can be marginal in some cases.  We make use of direct auto-differentiation in this work to encourage our techniques to be "plug-and-play" for neural network practitioners.  Lastly, and similarly to our work, \citet{basis-path} proposes improving on the 1-path-norm's bound by examining a subset of paths.  They identify a group of linearly independent paths and regularize using the resultant basis-path-norm in ReLU networks.


\section{Conclusion}

Our results demonstrate that optimization with 1-path-norm regularization is feasible in overparameterized neural networks using weight normalization and serves as a preferential method to achieve capacity control compared to standard practices.   We observe that regularization using our improved bound for our CReLU residual networks provides meaningful improvements in generalization.  

We confirm that both $L_1$ WN and 1-path-norm regularization synergistically contribute to producing near-sparsity effects and confirm that our weight pruning strategy is effective, although we expect alternative pruning strategies to work just as well.  Combining these features with length parameter sharing, PSiLON Net proves to be a competitive, reliable model, often outperforming random forests in tabular data.  Additionally, we observe that PSiLON ResNet's simplified regularization term grants it a small boost in convergence rate, while providing a meaningful reduction in compute.  In the small data regime, these models provide encouraging results for the practical use of overparameterized neural networks, where significant capacity control may be required.

Our work does not address the interaction of 1-path-norm regularization with other inductive biases such as convolutional layers, which is a limitation.  There is evidence $L_2$ weight regularization has a beneficial optimization effect, rather than a capacity control effect, when interacting with normalization layers \cite{l2-reg-vs-bn-and-wn}.  We leave exploring the benefits of combining $L_2$ weight regularization with 1-path-norm regularization to future work.

\section*{Impact Statement}
This paper presents work that may encourage the broader use of neural networks in the tabular and small data regimes, which are omnipresent in industry and fields such as medicine and economics.  There are many potential societal consequences of our work, none which we feel must be specifically highlighted here.

\bibliography{references}
\bibliographystyle{icml2024}

\newpage
\appendix
\onecolumn
\section{Orthogonal Projection onto the $L_1$ Sphere} \label{L1 proj appendix}
To perform an orthogonal projection onto the $L_1$ sphere, we may use Algorithm 1 in accordance with Proposition 2.1 given in \citet{fastprojection} with minor modification to ensure we land on the sphere rather than the ball.  This is achieved by enforcing the sign function to be nonzero and only matters in practice if the $L_1$ norm of the vector is in the interior of the ball and the vector has zero entries. This scenario is unlikely to begin with given the increasing norm property of $L_1$ and $L_2$ WN.  

This algorithm is amenable to vectorization and can be implemented in an auto-differentiation library for fast, automatic subgradients.  We reproduce a vectorized version of their (modified) algorithm in Algorithms 1 and 2.   The non-subdifferentiable steps include sorting, indexing, and the sign function.  As is standard in these libraries, sorting and indexing will simply route the gradient information, and the output of the sign function will be treated as a ``detatched" vector, simply serving to restore the information lost by the absolute value in Algorithm 2.  

\begin{algorithm}[h!]
   \caption{find\_threshold}
   \label{find threshold}
\begin{algorithmic}
   \STATE {\bfseries Input:} vector $\mathbf{w}$
   \STATE $\mathbf{u} \gets \text{sort}(\lvert \mathbf{w} \rvert, \text{descending=True})$
   \STATE $\mathbf{v} \gets (\text{cumsum}(\mathbf{u})-1) / (1:\text{length}(\mathbf{u}))$
   \STATE $i \gets \text{sum}(\text{cast\_to\_integer}(\mathbf{v} < \mathbf{u}))$
   \STATE $\tau \gets \mathbf{v}[i]$
   \STATE {\bfseries return} $\tau$
\end{algorithmic}
\end{algorithm}

\begin{algorithm}[h!]
   \caption{Orthogonal Projection onto the $L_1$ Sphere}
   \label{orthogonal project}
\begin{algorithmic}
   \STATE {\bfseries Input:} vector $\mathbf{w}$
   \STATE $\epsilon \gets 1\text{e-}8$
   \STATE $\tau \gets \text{find\_threshold}(\mathbf{w})$
   \STATE $\mathbf{p} \gets \text{max}(0, \lvert \mathbf{w} \rvert - \tau) \cdot \text{sign}(\mathbf{w} + \epsilon)$
   \STATE {\bfseries return} $\mathbf{p}$
\end{algorithmic}
\end{algorithm}

To make this strategy work with our CReLU residual networks, a minor modification is needed and is presented in Algorithm 3.  This ensures that $\mathbf{\tilde{w}}$ is projected onto the $L_1$ sphere as needed but allows the resulting projection for $\mathbf{w}^+$ and $\mathbf{w}^-$ to lie inside the $L_1$ ball.

\begin{algorithm}[h!]
   \caption{Orthogonal Projection into the $L_1$ Ball for CReLU Residual Network Weights}
   \label{orthogonal project crelu}
\begin{algorithmic}
   \STATE {\bfseries Input:} vectors $\mathbf{w}^+, \mathbf{w}^-$
    \STATE $\epsilon \gets 1\text{e-}8$
   \STATE $\mathbf{\tilde{w}} \gets \text{max}(\lvert \mathbf{w}^+ \rvert, \lvert \mathbf{w}^- \rvert)$
   \STATE $\tau \gets \text{find\_threshold}(\mathbf{\tilde{w}})$
   \STATE $\mathbf{p}^+ \gets \text{max}(0, \lvert \mathbf{w}^+ \rvert - \tau) \cdot \text{sign}(\mathbf{w}^+ + \epsilon)$
  \STATE $\mathbf{p}^- \gets \text{max}(0, \lvert \mathbf{w}^- \rvert - \tau) \cdot \text{sign}(\mathbf{w}^- + \epsilon)$
   \STATE {\bfseries return} $(\mathbf{p}^+, \mathbf{p}^-)$
\end{algorithmic}
\end{algorithm}

\section{Proof of Theorem 2} \label{proof appendix}
For a CReLU residual network $f_\mathbf{W}$, let $P_1(\mathbf{W})$ compute the 1-path norm when skip paths and weight paths are treated as distinct, i.e. we make use of a slightly worse upper bound resulting from $\lvert \mathbf{I}+\mathbf{W}^+_k \rvert \leq \mathbf{I}+ \lvert \mathbf{W}^+_k \rvert$, where the inequality holds elementwise, and similarly for $\mathbf{W}^-_k$.  Let $\tilde{\mathbf{W}}_k = \text{max}(\lvert \mathbf{W}^+_k \rvert, \lvert \mathbf{W}^-_k \rvert)$ and define
\begin{equation*}
	\tilde{P}_1(\mathbf{W}) =  \mathbf{1}^T \tilde{\mathbf{W}}_K \big( \mathbf{I} + \tilde{\mathbf{W}}_{K-1}  \big) \ldots \big( \mathbf{I} + \tilde{\mathbf{W}}_2 \big) \lvert \mathbf{W}_1 \rvert \mathbf{1}.
\end{equation*} 
If $\mathcal{L}_\mathbf{W}$ is the Lipschitz constant with respect to $L_\infty$ and $L_1$ for the input and output space, respectively, then we claim
\begin{equation*}
	\mathcal{L}_\mathbf{W} \leq \tilde{P}_1(\mathbf{W}) \leq P_1(\mathbf{W}).
\end{equation*}

The second inequality is obvious as we are summing over a strict subset of paths considered in $P_1(\mathbf{W})$. 

For the first inequality, we begin by assuming the output dimension of $\mathbf{W}_K$ is 1.  Let $\mathbf{z}^{(k)}$ be the hidden representation at layer $k$ and $\mathbf{z}^{(1)} = \mathbf{W}_1 \mathbf{x}$. Denote $\sigma^{+'}$ and $\sigma^{-'}$ as subgradients with respect to their arguments. With a slight abuse of notation, the output of the $(k+1)$-th layer of a CReLU newtork and its sub-Jacobian are given by
\begin{align*}
\mathbf{z}^{(k+1)} = f^{(k)}_\mathbf{W} ( \mathbf{x} ) &= \mathbf{z}^{(k)} + \mathbf{W}^+_k \sigma^+(\mathbf{z}^{(k)}) + \mathbf{W}^-_k \sigma^-(\mathbf{z}^{(k)})  \\
\nabla_{\mathbf{z}^{(k)}}  f^{(k)}_\mathbf{W} ( \mathbf{x} ) &= \mathbf{I} + \mathbf{W}^+_k \text{diag} \big( \sigma^{+'}(\mathbf{z}^{(k)}) \big) + \mathbf{W}^-_k \text{diag} \big( \sigma^{-'}(\mathbf{z}^{(k)}) \big).
\end{align*}
To compute
\begin{equation*}
	\mathcal{L}_\mathbf{W} = \sup_\mathbf{x} \lVert \nabla_\mathbf{x} f_\mathbf{W}(\mathbf{x})\rVert_{1} = \sup_\mathbf{x} \sup_{\lVert t \rVert_\infty \leq 1} D_\mathbf{x} f_\mathbf{W}(\mathbf{x}) t,
\end{equation*}
where $D_\mathbf{x}$ is a differential operator, we must use the chain rule and the previous expression.  Let $q(\mathbf{x})$ and $Q(\mathbf{x})$ be some vector and matrix, respectively, as functions of the input with dimensions compatible with their context. Observe that complementary bit vectors produced by $\sigma^{+'}$ and $\sigma^{-'}$ gives us
\begin{align*}
	&\sup_{\mathbf{x}} \sup_{\lVert t \rVert_\infty \leq 1} q(\mathbf{x})^T \big( \nabla_{\mathbf{z}^{(k)}}  f^{(k)}_\mathbf{W} ( \mathbf{x} )\big) Q(\mathbf{x}) t \\
	 \leq & \sup_{\mathbf{x}} \sup_{\mathbf{z}^{(k)}} \sup_{\lVert t \rVert_\infty \leq 1} q(\mathbf{x})^T \Big(\mathbf{I} + \mathbf{W}^+_k \text{diag} \big( \sigma^{+'}(\mathbf{z}^{(k)}) \big) + \mathbf{W}^-_k \text{diag} \big( \sigma^{-'}(\mathbf{z}^{(k)}) \big)  \Big) Q(\mathbf{x}) t  \\
	\leq & \sup_{\mathbf{x}} \sup_{\lVert t \rVert_\infty \leq 1}  \lvert q(\mathbf{x})^T \rvert \big( \mathbf{I} + \tilde{\mathbf{W}}_k \big)  \lvert Q(\mathbf{x}) \rvert t.
	\end{align*}
Assume the subgradient of $\sigma_\text{out}$ is globally bounded between zero and one. If we interpret $q(\mathbf{x})$ and $Q(\mathbf{x})$ as containing the remaining expressions in the chain rule expansion, we can apply these inequalities one residual block at a time (with trivial modification for the first and last linear maps as well as the final nonlinearity) to give us
\begin{align*}
	\mathcal{L}_\mathbf{W} &\leq \sup_{\lVert t \rVert_\infty \leq 1} \tilde{\mathbf{W}}_K \big( \mathbf{I} + \tilde{\mathbf{W}}_{K-1}  \big) \ldots \big( \mathbf{I} + \tilde{\mathbf{W}}_2 \big) \lvert \mathbf{W}_1 \rvert t \\
	&\leq \tilde{\mathbf{W}}_K \big( \mathbf{I} + \tilde{\mathbf{W}}_{K-1}  \big) \ldots \big( \mathbf{I} + \tilde{\mathbf{W}}_2 \big) \lvert \mathbf{W}_1 \rvert \mathbf{1}.
\end{align*}
To account for arbitrary output dimension, note that since we can additively decompose the $L_1$ norm by each output dimension, it was sufficient to show the inequality for output size 1.  By simply summing the inequalities associated with each output dimension, we get $\tilde{P}(\mathbf{W})$ and the desired result.  This approach is similar to Lemma 3 of \citet{1pathnorm-dnn}, which they use to prove their Theorem 1 (reproduced and numbered as Theorem 1 in our work as well).

\section{Neural Network Training Details}
\subsection{Small Tabular Datasets} \label{small tabular appendix}
We use orthogonal initialization for all weight matrices and initialize all length parameters used in normalization to 1. Both models are given the same initialization. 

We use 5 batches per epoch, leading to a batch size of 400.   Our optimization objective is the regularized cross-entropy loss for classification problems and the regularized mean-squared-error loss for regression problems.  The Adam optimizer is used with default parameters except for the learning rate. We schedule the learning rate as follows.  We linearly increase the learning rate from 1e-4 to 2e-3 for the first 5\% of training iterations, maintain this learning rate for the next 45\%, and linearly decrease the learning rate back down to 1e-4 during the final 50\%. 

For our hyperparameter search, we choose the regularization hyperparameter that leads to the best validation loss from the following set: $\{5\text{e-}5, 1\text{e-}4, 2.5\text{e-}4, 5\text{e-}4, 1\text{e-}3, 2.5\text{e-}3, 5\text{e-}3, 1\text{e-}2, 2.5\text{e-}2, 5\text{e-}2, 1\text{e-}1, 2.5\text{e-}1, 5\text{e-}1 \}$. This set applies to both P-Net and S-Net. No hyperparameters were chosen at the boundaries of this set for any model. For our random forest models, we fix the number of trees at 1000 and constrain the minimum number of samples allowed in a leaf.  We choose this hyperparameter from the following set: $\{4,8,12,16,20,30,40,50,75,100,150,200\}$.

\subsection{Ablation Study on Deep Networks} \label{ablation deep appendix}
We use orthogonal initialization for all weight matrices.  We set each $\mathbf{W}^-_k$ matrix's initialization to be equal to its corresponding $\mathbf{W}^+_k$ matrix to create a ``looks linear" initialization.  Length parameters are initialized to 1 for the first and last linear maps; they are initialized to 0 for all interior linear maps used in the residual layers. All models are given the same initialization.

We use 50 batches per epoch, leading to a batch size of 200.  Our optimization objective is the regularized cross-entropy loss.  The Adam optimizer is used with default parameters except for the learning rate.  We schedule the learning rate according to the OneCycle strategy \cite{onecycle}, where we use a maximum learning rate of $2\text{e-}2$ achieved 20\% into training as well as initial and final learning rates of $1\text{e-}4$ and $1\text{e-}5$, respectively.

For our hyperparameter search, we choose the regularization hyperparameter that leads to the best validation loss from the following set for all models except S-ResNet: $\{1\text{e-}5, 2.5\text{e-}5, 5\text{e-}5, 1\text{e-}4, 2.5\text{e-}4, 5\text{e-}4, 1\text{e-}3 \}$.  As S-ResNet uses a different regularization term that produces smaller values, we search for its hyperparameter within $\{1\text{e-}3, 2.5\text{e-}3, 5\text{e-}3, 1\text{e-}2, 2.5\text{e-}2, 5\text{e-}2, 1\text{e-}1 \}$.  No hyperparameters were chosen at the boundaries of these sets for any models.

\end{document}